\documentclass[a4paper,fleqn]{arxiv}
\usepackage{graphicx} 
\usepackage{float} 
\usepackage{subcaption}
\usepackage{tcolorbox}
\usepackage{tabularray}
\UseTblrLibrary{siunitx}
\usepackage{array}
\usepackage{makecell}
\usepackage{tabularx}
\usepackage{colortbl}
\usepackage{hhline}   
\usepackage{multirow} 
\usepackage{booktabs}
\usepackage{amsmath}  
\usepackage{amsfonts}   
\usepackage{amssymb} 
\usepackage{mathtools}
\usepackage{esint,upgreek}
\usepackage{subfiles}
\usepackage{blindtext}
\usepackage{minitoc}
\usepackage{setspace}
\usepackage{endnotes} 
\usepackage{fancyvrb} 
\usepackage{rotating}
\usepackage{xurl}

\usepackage{mylines}
\usepackage{myplots}
\usepackage[numbers]{natbib}

\definecolor{training_color}{HTML}{ee6C4d} 
\definecolor{testing_color}{HTML}{293241} 
\definecolor{gray_plane}{HTML}{787878} 

\definecolor{quiver_ref_color}{HTML}{89bedc} 
\definecolor{quiver_pred_color}{HTML}{084387} 
\definecolor{quiver_arrow_color}{HTML}{ee6C4d} 

\definecolor{no_pca_color}{HTML}{ee6C4d} 
\definecolor{pca_color}{HTML}{293241} 

\definecolor{cw_ref_color}{HTML}{000000} 
\definecolor{cw_pred_color}{HTML}{e32f27} 

\definecolor{ae_error_color}{HTML}{98d594} 
\definecolor{full_model_error_color}{HTML}{ee6c4d} 
\definecolor{full_ft_model_error_color}{HTML}{89bedc} 
\providecommand{\norm}[1]{\lVert#1\rVert} 

\def\tsc#1{\csdef{#1}{\textsc{\lowercase{#1}}\xspace}}
\tsc{WGM}
\tsc{QE}

\begin{document}
\let\WriteBookmarks\relax
\def\floatpagepagefraction{1}
\def\textpagefraction{.001}

\shorttitle{Towards aerodynamic surrogate modeling based on $\beta$-variational autoencoders}

\shortauthors{V. Franc\'es-Belda et al.}  

\title [mode = title]{Towards aerodynamic surrogate modeling based on $\beta$-variational autoencoders}

\affiliation[1]{organization={Theoretical and Computational Aerodynamics Branch, Spanish National Institute for Aerospace Technology (INTA)},
            city={Torrej\'on de Ardoz},
            postcode={28850}, 
            state={Madrid},
            country={Spain}}
\affiliation[2]{organization={Subdirectorate General of Terrestrial Systems, Spanish National Institute for Aerospace Technology (INTA)},
            city={San Mart\'in de la Vega},
            postcode={28330}, 
            state={Madrid},
            country={Spain}}            
\affiliation[3]{organization={Department of Aerospace Engineering, Universidad Carlos III de Madrid},
            city={Legan\'es},
            postcode={28911}, 
            state={Madrid},
            country={Spain}}

\author[1]{V\'ictor Franc\'es-Belda}[
    orcid=0009-0003-7083-1520]
\ead{vfrabel@inta.es}
\author[2,3]{Alberto Solera-Rico}[
    orcid=0000-0003-3883-5989]
\ead{asolric@inta.es}
\author[1,3]{Javier Nieto-Centenero}[
    orcid=0000-0003-2764-6082]
\ead{jniecen@inta.es}
\author[1]{Esther Andr\'es}[
    orcid=0000-0003-4559-4730]
\ead{eandper@inta.es}
\author[2,3]{Carlos Sanmiguel Vila}[
    orcid=0000-0003-0331-2854]
\cormark[1] 
\ead{csanvil@inta.es}

\author[3,2,1]{Rodrigo Castellanos}[
    orcid=0000-0002-7789-5725]
\cormark[1] 
\ead{rcastell@ing.uc3m.es}
\cortext[1]{Corresponding authors}

\maketitle

\noindent \textsc{Abstract} \\
\\
Surrogate models combining dimensionality reduction and regression techniques are essential to reduce the need for costly high-fidelity CFD data. New approaches using $\beta$-Variational Autoencoder ($\beta$-VAE) architectures have shown promise in obtaining high-quality low-dimensional representations of high-dimensional flow data while enabling physical interpretation of their latent spaces. We propose a surrogate model based on latent space regression to predict pressure distributions on a transonic wing given the flight conditions: Mach number and angle of attack. The $\beta$-VAE model, enhanced with Principal Component Analysis (PCA), maps high-dimensional data to a low-dimensional latent space, showing a direct correlation with flight conditions. Regularization through $\beta$ requires careful tuning to improve the overall performance, while PCA pre-processing aids in constructing an effective latent space, improving autoencoder training and performance. Gaussian Process Regression is used to predict latent space variables from flight conditions, showing robust behavior independent of $\beta$, and the decoder reconstructs the high-dimensional pressure field data. This pipeline provides insight into unexplored flight conditions. Additionally, a fine-tuning process of the decoder further refines the model, reducing dependency on $\beta$ and enhancing accuracy. The structured latent space, robust regression performance, and significant improvements from fine-tuning collectively create a highly accurate and efficient surrogate model. Our methodology demonstrates the effectiveness of $\beta$-VAEs for aerodynamic surrogate modeling, offering a rapid, cost-effective, and reliable alternative for aerodynamic data prediction.

\vspace{1cm}

\section{Introduction}
Obtaining reliable aerodynamic models remains a significant challenge in aeronautics, particularly during the aircraft design phase, which demands comprehensive performance characterization across various flight conditions \cite{fielding2017introduction, Li2022}. Advanced Computational Fluid Dynamics (CFD) techniques provide high-fidelity data, but they are computationally intensive and impractical for iterative design processes. This limitation has driven interest in developing surrogate models that approximate aerodynamic behavior at a fraction of the computational cost. However, creating accurate surrogate models is difficult due to the requirement for high-quality training data and the ability to generalize across diverse conditions \cite{yondo2018review, zhang2021multi, yondo2019review}. 

Surrogate models simulate aerodynamic characteristics without requiring extensive CFD simulations by utilizing data from high-fidelity simulations or experiments and applying various mathematical and statistical techniques \cite{forrester2008engineering}. These models facilitate rapid iteration and optimization during the design process by significantly reducing computational resources and time \cite{du2021rapid, du2022airfoil}.
A common approach to building aerodynamic surrogate models involves using low-dimensional reduction techniques followed by regression modeling. Dimensionality reduction methods are first applied to high-dimensional aerodynamic data to capture the most significant features with fewer dimensions. The reduced-dimensional data is then used to train regression models, establishing a relationship between the simplified input space and the aerodynamic outputs.  Among the regression models explored in aerodynamic surrogates, some options are K-Nearest Neighbors (kNN) \cite{Wang2019knn}, Gaussian Process Regression (GPR) \cite{NietoCentenero2023fusing}, polynomial chaos kriging \cite{Zhao2022pck}, and architectures based on Neural Networks (NN) \cite{Immordino2024dnn}, among others.

For dimensionality reduction, techniques like Proper Orthogonal Decomposition (POD) 
and Principal Component Analysis (PCA) 
are commonly used. POD is used for reduced-order modeling of complex problems \cite{berkooz1993proper}, within a Galerkin projection framework (POD-Galerkin) \cite{Lucia2004pod}, combined with a CFD flux residual minimization scheme \cite{zimmermann2012pod_cfd}, or coupled with an interpolation method \cite{Ly2001PODI, Bui2004pod}. The use of POD (or PCA) in surrogate modeling is extensively explored \cite{iuliano2013proper, dolci2016proper, dupuis2018surrogate}; however, this linear technique has limitations in capturing non-linear complexities. Although POD has been applied to nonlinear problems, its underlying assumption that the input data lie in a low-dimensional, linear subspace makes it insufficient for highly nonlinear features. In aerodynamics, accurately predicting shockwaves in the transonic flow regime poses a severe challenge for POD-based methods because such flows result in sharp, discontinuous changes in flow variables like pressure, temperature, density, and velocity across the shockwave.

To address the limitations of linear assumptions in POD, manifold learning techniques offer a viable alternative. Manifold learning, a branch of statistical learning, aims to recognize the topologically closed surface (the manifold) over which the data lie on or near it \cite{meilua2024manifold}. The manifold is a geometric representation of the intrinsic relationships that connect data samples. Examples include Isomap, 
Multi-dimensional Scaling (MDS), 
and Local Linear Embedding (LLE). 
Surprisingly, the application of manifold learning in aeronautics is not widely exploited. \citet{franz2014interpolation} combined, for the first time, Isomap with a thin plate spline regressor into a surrogate model to predict the pressure distribution ($C_p$) on a wing in the transonic regime. Similarly, \citet{castellanos2022assessment} combined Isomap with a NN, improving the surrogate's performance in predicting shockwaves. Isomap has also been used for adaptive sampling \cite{halder2024adaptive}, leading to a method that iteratively generates samples based on a low-dimensional manifold of the training data. Recently, LLE has been applied to develop a multi-fidelity surrogate model for efficient aerodynamic predictions \cite{NietoCentenero2024lle}. The challenge of these techniques is that they require additional regression algorithms to decode the low-dimensional manifold back to the original dimension of the input data, adding further complexity to the surrogate \cite{Castellanos2023isomap}. Other alternatives to classic POD (or PCA) and manifold learning strategies include Gauss-Newton with Approximated Tensors (GNAT), a projection-based framework related to residual minimization \cite{Carlberg2013GNAT}, or the Discrete Empirical Interpolation Method (DEIM) \cite{Chaturantabut2010DEI}, which rely on a classical POD-Galerkin approach while allowing the evaluation of nonlinear terms of the governing equations with an additional POD basis. However, their application to aerodynamic surrogate modeling still needs to be well established.

Recent advances in neural networks provide a promising framework for developing surrogate models that address the limitations of previous approaches \cite{du2021rapid, liao2021multi, du2022airfoil, duru2022deep, deng2023prediction, wang2023general, Hines2023mlp}. Aerodynamic surrogates based on graph neural networks (GNNs) have gained interest due to their ability to efficiently handle unstructured data and focus on local features, as demonstrated by \citet{Hines2023gnn} in predicting aircraft surface pressure distributions. In addition, neural networks are also employed for dimensionality reduction. In particular, autoencoders (AE) \cite{wang2021flow, kang2022physics, wang2023physics, beiki2023novel} have been increasingly used in aerodynamic modeling, effectively addressing the limitations of POD and manifold learning by non-linear transformations of the data into compact low-dimensional spaces while providing a global encoding-decoding mechanism.

Among the different AE architectures, Variational Autoencoders (VAE) \cite{kingma2013auto} have been proven to be effective in encoding the spatial information of fluid flows into low-dimensional non-linear latent spaces \cite{eivazi2020deep, zhang2023nonlinear}. Unlike traditional AEs, VAEs incorporate a probabilistic approach for representing observations in the latent space by adding an extra loss term specific to the latent variables. This loss term allows VAEs to learn a more continuous latent space that better fits posterior regressor models. However, a notable limitation of these models is the lack of orthogonality, a feature inherent in classical linear decomposition techniques such as POD, which can result in several entangled features within a single latent variable, complicating the interpretation of the latent space.
To address this limitation, the $\beta$-Variational Autoencoder ($\beta$-VAE) architecture \cite{higgins2017betavae} has been proposed. This architecture modifies the VAE's loss function by incorporating a regularization parameter, $\beta$, which balances reconstruction accuracy with regularization and latent space disentanglement \cite{burgess2018understanding}. The $\beta$ parameter is carefully selected to ensure that the latent space representation is nearly orthogonal while minimizing any increase in reconstruction error. The potential of these architectures is outlined in various studies where different Reduced Order Models (ROMs) were developed using this architecture to describe the evolution of turbulent flow fields \cite{eivazi2022vae, solera2024beta, wang2024towards}. These studies report excellent performance in reconstructing turbulent flows and have found that the latent space features in these architectures represent the most energetic POD modes. These recent findings suggest that $\beta$ -VAE can capture nonlinear modes that represent the most energetic or representative flow characteristics \cite{solera2024beta, wang2024towards}. Consequently, these architectures hold significant potential for developing physically interpretable surrogate models \cite{kang2022physics}.

Notable applications of AE-based architectures in surrogate modeling for aerodynamics include the recent use of graph convolutional autoencoders for steady-state aerodynamics of transonic aircraft \cite{Massegur2024gcae, immordino2024gcae}, although they can be extended to other fields such as aeroacoustic airfoil shape optimization \cite{kou2023aeroacoustic} or modeling three-dimensional unsteady flows \cite{Gupta2022dlrom}.
Specifically for $\beta$-VAE models, \citet{kang2022physics} present a study of the extraction of information-intensive latent variables (LVs) from two-dimensional airfoil flow data and explores how these low-dimensional LVs correlate with the physical properties of the flow. This work illustrates the ability of the $\beta$-VAE models to discard redundant LVs as a result of proper tuning of the $\beta$ hyperparameter, and the argument is made that this behavior is indicative of the physical awareness of the resulting model.
In a related approach, \citet{moni2024deimvae} used DEIM to non-linearly derive a reduced set of variables from flows around airfoils and wings. The reduced set of variables was then used to train an autoencoder with clustering (AE-C), to obtain a disentangled low-dimensional representation of the data with a prescribed number of clusters. In the clustered space, a nearest-neighbor algorithm is used to select the nearest point, which is then used to reconstruct the flow field prediction.

Due to the potential of these methods and the need to further analyze their performance in 3D flows, this study leverages the physically interpretable and information-rich LVs derived from $\beta$-VAEs for efficient surrogate modeling of aerodynamic pressure distribution. Using a comprehensive database of CFD simulations, we demonstrate the ability of $\beta$-VAEs to extract key aerodynamic features autonomously. Our proposed surrogate modeling framework integrates $\beta$-VAEs with PCA preprocessing to compress high-dimensional input data without loss of information. This novel approach involves a two-step reduction in dimensionality, leading to more efficient $\beta$-VAE training. Then, GPR maps the physical ROM input variables to the $\beta$-VAE latent space points. Finally, this work includes a final fine-tuning process that reduces the dependence on the $\beta$ hyperparameter, allowing easier model tuning. This integration and the fine-tuning process improve the accuracy and generalization capability of the model, marking a significant advancement in aerodynamic surrogate modeling using autoencoders.

The remainder of this paper is organized as follows: \autoref{s:database} outlines the CFD database and test case. Section \ref{s:methodology} describes the detailed methodology, including the $\beta$-VAE framework (\S \ref{ss:bVAE}), GPR (\S \ref{ss:GPR}), PCA pre-processing (\S \ref{ss:PCA}), and description of the surrogate model (\S \ref{ss:surrogate}). The analysis of the latent space, discussing the physical interpretability of the LVs, in \autoref{s:latent}, is followed by the performance assessment of the surrogate model in \autoref{s:results}, where a novel approach for tuning the decoder is proposed (\S \ref{ss:finetunning}). Finally, \autoref{s:conclusions} outlines the conclusions of the study.

\section{Database and test case: XRF1 wing}\label{s:database}
The selected test case comprises a database of CFD simulations of the XRF1 model. This model, provided by Airbus{\texttrademark}, is a research test case to showcase various technologies applied to long-range widebody aircrafts. The database was developed within the AD/AG60 research project of the Group for Aeronautical Research and Technology in Europe (GARTEUR). Further information on the XRF1 research model can be found in \citet{Pattinson2013xrf1}.

\begin{figure}[tb]
    \centering
    \includegraphics[width=0.99\linewidth]{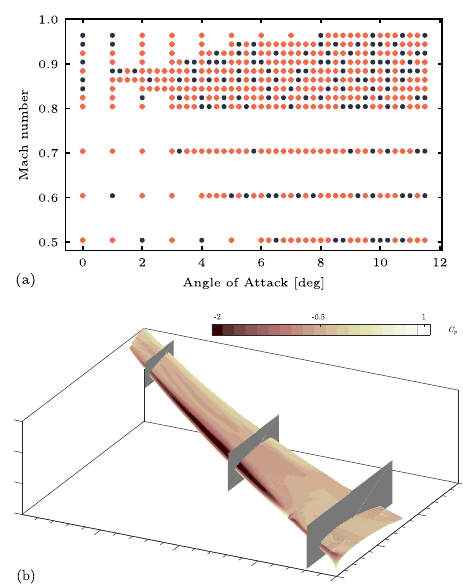}
    \caption{(a) Flight envelope with the tested flight conditions split in \textit{training} set \sy{training_color}{d*} and \textit{test} set \sy{testing_color}{o*}. (b) Example of $C_p$ distribution over the wing for $\left(\alpha,M\right)=\left(9.25^\circ, 0.7\right)$. Three planes are highlighted at wingspan percentages $\eta=0.1, 0.5, 0.9$ \lcap{_}{gray_plane}. Note that the wing deformation is the same for all the cases.}
    \label{fig:dataset_wing}
\end{figure}

Aerodynamic data were derived from Reynolds Averaged Navier-Stokes (RANS) simulations of the complete aircraft model to ensure realistic conditions through the interaction of various aerodynamic subsystems. The dataset focuses on the dimensionless pressure coefficient ($C_p$) distribution over the wing for different flight conditions (see an example in \autoref{fig:dataset_wing}b). It comprises 435 distinct flight conditions computed using the DLR TAU solver \cite{Kroll2014tau} at a fixed Reynolds number ($Re = 2.5 \times 10^7$). The flight conditions span the entire flight envelope of the aircraft, with Mach numbers ($M$) ranging from 0.5 to 0.96 and angles of attack ($\alpha$) from $0^\circ$ to $11.5^\circ$ (see \autoref{fig:dataset_wing}a).

The 435 flow solutions in the database are randomly split into two sets: a \textit{training} set (70\% of the database, 305 cases) used to train the surrogate model [\sy{training_color}{d*} in \autoref{fig:dataset_wing}a], and a \textit{test} set (30\% of the database, 130 cases) to evaluate the model's ability to predict $C_p$ within the entire design space [\sy{testing_color}{o*} in \autoref{fig:dataset_wing}a]. The cases in the \textit{test} set are excluded from the training process. To visualize the performance of the models, three cases from the \textit{test} set are chosen: low complexity $\left(\alpha_1,M_1\right)=\left(2.0^\circ, 0.504\right)$, intermediate complexity $\left(\alpha_2,M_2\right)=\left(5.75^\circ, 0.704\right)$, and high complexity $\left(\alpha_3,M_3\right)=\left(9.75^\circ, 0.904\right)$, the latter featuring shockwaves on the upper wing surface. Although the performance of the model will be evaluated in the whole \textit{test} set, these three cases will be used to visualize the error upon prediction with the proposed models.

The fixed load factor $n_z$ ($= L/W$, where $L$ is the lift and $W$ is the weight) of 2.5 poses an additional challenge for the prediction model due to significant wing deformation under high loads, as can be seen in \autoref{fig:dataset_wing}b. This condition represents the limit load factor for transport category aircraft per airworthiness regulations. The wing undergoes structural deformation at such conditions but returns to its normal shape once the load factor returns to $n_z = 1$ (typical cruise condition). Aerodynamically, the deflection of the wing geometry prevents shockwave formation on the upper surface, though significant $C_p$ changes and separation regions occur within the flight envelope. This condition is critical for surrogate model development due to its relevance to certification requirements.

Based on the load factor condition, we only focus on the upper wing surface since the lower surface does not challenge the predictive models, as previously seen in \citet{castellanos2022assessment}.  Additionally, discarding the lower surface avoids the numerical artifacts in the model due to irregularities in the mesh around the engine pylon. Consequently, each $C_p$ sample is defined on an unstructured mesh of $q = 49574$ grid points. Each $C_p$ distribution is an observation (point) in the high-dimensional space $\mathbb{R}^q$, where each dimension (feature) corresponds to a $C_p$ point. Let $\mathbf{X}\in{\mathbb R}^{q\times n}$ represent the data matrix, where $n$ indicates the number of available flight conditions (that is, 435). Each column vector, $\mathbf{x}_i\in \mathbb{R}^q$, corresponds to $C_{p,i}$ for $i=1,\ldots,n$. The dataset in $\mathbf{X}$ is inherently complex, challenging the extraction of a reduced set of meaningful coordinates that capture the main flow characteristics. Hereafter, data samples are denoted as $\mathbf{x}$, the dataset as $\mathbf{X}$, autoencoder reconstructions as $\mathbf{\tilde{x}}$, and surrogate model predictions as $\mathbf{\hat{x}}$. For clarity in the figures, the pressure distribution coefficient and its reconstructions and predictions are denoted as $C_p$, $\tilde{C}_p$, and $\hat{C}_p$, respectively.

\section{Surrogate Modeling Framework and Methods} \label{s:methodology}
The application of Machine Learning (ML) techniques focused on generating surrogate models has been in the spotlight of many engineering fields due to their capacity to relieve the computational costs associated with CFD simulations. This section is devoted to reviewing the different techniques that this work follows to build the surrogate.

\subsection{$\beta-$variational autoencoder} \label{ss:bVAE}
Autoencoders are a recent field of research focused on using neural networks to derive meaningful, low-dimensional representations from high-dimensional data, leveraging the nonlinear nature of neural networks and their versatility in handling various types of data. They are an unsupervised learning strategy that aims to learn a compressed representation (latent space) of the training data by encoding it into a lower-dimensional space through an encoder network and, then reconstructing the input from this latent representation using a decoder network.

We denote as $\mathbf{z}_i$ the $i$-th component of $\mathbf{Z}\in\mathcal{Z}$, being $\mathcal{Z}\subset \mathbb{R}^d$ the latent space of dimensionality $d$, which is significantly smaller than the input dimension ($d\ll q$). The encoder network maps the input data, $\mathbf{x}$ in the data space $\mathcal{X}\subset \mathbb{R}^{q \times n}$, to the latent space ($E:\mathcal{X}\rightarrow\mathcal{Z}$) and the decoder network reconstructs the input data from the latent representation ($D:\mathcal{Z}\rightarrow\mathcal{X}$). The data is compressed into the latent space in the bottleneck between the encoder and decoder networks. Let $\tilde{\mathbf{x}} = (D\circ E)(\mathbf{x})$ be the output of the autoencoder, which is trained to minimize a reconstruction loss function $\mathcal{L}_{rec}$, to drive the reconstructed output data to match the input data. The standard autoencoder loss is given by the mean squared error (MSE),
\begin{equation}
    \mathcal{L}_{rec} = \frac{1}{q}\norm{ \mathbf{x} - (D\circ E)(\mathbf{x}) }_2^2 ,
\end{equation}
where $\norm{\bullet}_2$ denotes the $\ell^2$-norm operator.

While traditional autoencoders are effective at capturing significant features of the input data, there is no explicit control over the properties of the underlying latent space. Furthermore, the lack of constraints in the latent space can lead to overfitting and poor generalization. VAEs \cite{kingma2013auto} address this limitation by introducing probabilistic modeling into the encoding process. They promote a more structured and continuous latent space by learning to generate latent representations that follow a predefined prior probability distribution, typically a Gaussian distribution. Therefore, the encoding process yields a probability distribution per each input point of dimension $q$ that is characterized by a mean $\boldsymbol{\mu}$ and a standard deviation $\boldsymbol{\sigma}$ of dimensionality $d$. In order to perform a correct optimization of the parameters of the network, the latent vectors are modeled following the so-called reparametrization trick, 
\begin{equation}
    \mathbf{z} = \boldsymbol{\mu} + \boldsymbol{\sigma} \odot \boldsymbol{\varepsilon}, \quad \boldsymbol{\varepsilon} \sim \mathcal{N}(0, 1), 
\end{equation}
where $\odot$ refers to the element-wise product, and $\mathcal{N}(0, 1)$ refers to a Gaussian distribution with a mean of 0 and standard deviation of 1.

The training loss includes an additional term: the Kullback--Leibler (KL) divergence loss~\cite{KullbackLeibler}, $\mathcal{L}_{KL}$, which quantifies the difference between the generated probability distributions and a prior probability distribution. Typically, a standard Gaussian distribution with mean $\mu=0$ and standard deviation $\sigma=1$ is assumed. Hence, a classic definition of the loss function for a VAE is
\begin{equation}
    \mathcal{L} (\mathbf{x}) = 
    \underbrace{\frac{1}{q}\norm{\mathbf{x}-
    \tilde{\mathbf{x}}}_2^2 \vphantom{\frac{1}{2}\sum^d_{i=1}\big(1+\log(\sigma_i^2)-\mu_i^2-\sigma_i^2 \big)}}_{\mathcal{L}_{rec}}
    - 
    \underbrace{\frac{1}{2} \sum^d_{i=1}\big(1+\log(\sigma_i^2)-\mu_i^2-\sigma_i^2 \big)}_{\mathcal{L}_{KL}}.
    \label{ec:VAEloss}
\end{equation}

The key innovation of VAEs is incorporating a variational inference framework, which enables the model to learn the mean and variance of the latent variables during training. Samples are taken from the distribution and decoded to the high-dimensional space. This stochastic method produces a more continuous latent space that allows VAEs to perform generative tasks by sampling new points in the latent space.

$\beta$-VAE architectures, as proposed in \citet{higgins2017betavae}, extend the VAE framework by introducing a hyperparameter $\beta$ that controls the trade-off between the reconstruction accuracy and the disentanglement of latent variables. Greater values of $\beta$ encourage more disentangled representations, which can be beneficial for tasks requiring semantic control over generated data, with the cost of increasing the reconstruction error,
\begin{equation}
    \mathcal{L}(\mathbf{x}) = \mathcal{L}_{rec} - \beta\mathcal{L}_{KL}.
    \label{ec:betaVAEloss}
\end{equation}

In this paper, using $\beta$-VAEs to map pressure distribution fields enables the model to leverage a nonlinear dimensionality reduction method with increased fidelity over a classical linear method.

\subsection{Gaussian Process Regression} \label{ss:GPR}
The surrogate model proposed in this study (see \autoref{ss:surrogate}) employs GPR to perform regression on variables in the latent space derived from the $\beta$-VAE. Using flight conditions as input data points, \(\mathbf{P} = \{\mathbf{p}_1, \mathbf{p}_2, \ldots, \mathbf{p}_n \}\in\mathcal{P}\) (the parameter space), where each \(\mathbf{p}_i = [M_i, \alpha_i]\), the GPR model predicts the set of \(d\) latent variables within the latent space, \(\mathbf{Y} = \{\mathbf{y}_1, \mathbf{y}_2, \ldots, \mathbf{y}_n \}\), where each \(\mathbf{y}_i = [\mu_{i1}, \ldots, \mu_{id}]\) represents the latent variables associated with \(\mathbf{p}_i\). The choice of GPR is justified by its effectiveness in performing nonlinear regressions, offering high accuracy and flexibility, thereby ensuring robust and reliable models. An overview of GPR theory is presented below. For a detailed mathematical analysis, readers are referred to comprehensive literature on the topic \cite{williams2006gaussian, gramacy2020surrogates, melo2012gaussian}.

A Gaussian process (GP) is a collection of random variables, any finite number with a joint Gaussian distribution. It is characterized by its mean function, $m(\mathbf{p})=\mathbb{E}[f(\mathbf{p})]$, and covariance function, $k(\mathbf{p},\mathbf{p}^{\prime}) = \mathbb{E}[(f(\mathbf{p}) - m(\mathbf{p})) (f(\mathbf{p}^{\prime})-m(\mathbf{p}^{\prime}))]$, where $\mathbf{p}$ is the input vector, and $\mathbb{E}$ is the expected value. Thus, the classic definition of a GP is given by:

\begin{equation}\label{eq:definition_GP}
    f(\mathbf{p})\sim \mathbb{GP} \left(m(\mathbf{p}),k(\mathbf{p},\mathbf{p}^{\prime})\right).
\end{equation}

Although not required, the mean function typically takes a zero value for simplicity of notation. Each element of the training dataset, $\mathbf{y}$, is a sample with a Gaussian distribution representing the true value of the observation $f(\mathbf{p})$, perturbed by some independent Gaussian noise $
\boldsymbol{\varepsilon}$, with variance $\sigma_{n}^2$. Therefore, the observations can be expressed as $\mathbf{y}=f(\mathbf{p})+\boldsymbol{\varepsilon}$. The goal of the regression is to predict $\mathbf{F}_*$ values at new points $\mathbf{P_*}$. The joint distribution of the training values and the function at new points is given by:
\begin{equation}
    \left[\begin{array}{c}
    \mathbf{Y} \\
    \mathbf{F}_*
    \end{array}\right] \sim \mathcal{N}\left( \mathbf{0},\left[\begin{array}{cc}
    K(\mathbf{P}, \mathbf{P}) + \sigma_n^2 \mathbf{I} & K\left(\mathbf{P}, \mathbf{P}_*\right) \\
    K\left(\mathbf{P}_*, \mathbf{P}\right) & K\left(\mathbf{P}_*, \mathbf{P}_*\right)
    \end{array}\right]\right) \; ,
\end{equation}
where $\mathbf{P}_* = \mathbf{[p_{*1},\ldots, p_{*n}]}$ are the new locations where the predictions will be made, and $\mathbf{K}$ are matrices constructed using any function $k(\mathbf{p,p^{\prime}})$ that generates a non-negative deﬁnite covariance matrix. These functions are known as kernel functions.
    
By deriving the conditional distribution, the predictive equations for GPR are derived as
\begin{equation} \label{eq:predictive_GPR}
\overline{\mathbf{F}}_*|\mathbf{P},\mathbf{Y},\mathbf{P}_*\ \sim \mathcal{N}(\overline{\mathbf{F}}_*,\ \operatorname{cov}(\mathbf{F}_*)),
\end{equation}
where 
\begin{equation} \label{eq:mean_GPR}
    \overline{\mathbf{F}}_* = K\left(\mathbf{P}_*, \mathbf{P}\right)\left[K(\mathbf{P}, \mathbf{P})+\sigma_n^2 \mathbf{I}\right]^{-1} \mathbf{Y} \;, 
\end{equation} 
and
\begin{align}\label{eq:cov_GPR}
    \operatorname{cov}(\mathbf{F}_*) = & \, K\left(\mathbf{P}_*, \mathbf{P}_*\right) \nonumber \\
    & - K\left(\mathbf{P}_*, \mathbf{P}\right) \left[K(\mathbf{P}, \mathbf{P}) + \sigma_n^2 \mathbf{I} \right]^{-1} K\left(\mathbf{P}, \mathbf{P}_*\right) \;.
\end{align}
            
Learning the noise variance and the kernel hyperparameters is achieved through the maximization of the log marginal likelihood given by
\begin{align} \label{eq:LFL_GPR}
    \log p(\mathbf{Y}|\mathbf{P}) = & -\frac{1}{2} \mathbf{Y}^{\mathsf{T}} [K(\mathbf{P}, \mathbf{P}) + \sigma^2 \mathbf{I}]^{-1} \mathbf{Y} \nonumber \\
    & - \frac{1}{2} \log |K(\mathbf{P}, \mathbf{P}) + \sigma^2 \mathbf{I}| - \frac{n}{2} \log(2\pi) \;.
\end{align}

The details on the features of the GPR model are described in \autoref{ss:surrogate}, defining the kernels under consideration. In addition to the kernel, the Automatic Relevance Determination (ARD) technique is also used. Unlike its conventional application, where ARD is used to identify and discard irrelevant features, in this particular case, where both inputs are relevant to the model, ARD allows for the individual adjustment of the input scales. This technique is implemented by adjusting a different kernel hyperparameter for each input, enabling the model to capture each variable's specific characteristics adequately. This approach results in a more accurate and robust representation of the relationships in the data without losing valuable information from any input variables.

\subsection{Data pre-processing stage} \label{ss:PCA}
Fully connected neural networks, as is the case of the herein considered $\beta-$VAE, most often contain a high number of trainable parameters, which can rapidly increase with high input dimensions. Additionally, training such large models requires a considerable number of data samples to ensure convergence. To mitigate these challenges, dimensionality reduction techniques can be applied before model training to reduce the initial size of the data, thus reducing computational costs \cite{Lee2023podvae}.

PCA is a major application of SVD, offering a statistical interpretation of a hierarchical coordinate system for high-dimensional correlated data. Such a coordinate system is determined by uncorrelated (orthogonal) \textit{principal components} that maximally correlate with the input data \cite{Pearson1901pca}, leading to a fully data-driven technique for reducing data dimensionality. We propose the use of PCA to reduce the dimensionality of the input data, allowing the architecture of the autoencoder to be relatively shallow, which means that the deep learning model is fast to train and, more importantly, does not need as much training data \citep{cantero2023data}. The PCA formulation is briefly introduced to clarify and confirm the nomenclature used in this manuscript.  Considering the data matrix $\mathbf{X}$ (as defined in \autoref{s:database}), classic PCA formulation requires pre-processing by subtracting the mean and setting the variance to unity before performing SVD on the covariance matrix $\mathbf{C}$. The row-wise zero empirical mean matrix $\mathbf{B} = \mathbf{X} -\overline{\mathbf{X}}$ is computed, where $\overline{\mathbf{X}}$ is the row-wise mean matrix of $\mathbf{X}$. Invoking Bessel's correction, the normalized, zero-empirical-mean covariance matrix is computed as $\mathbf{C} = \mathbf{B}^*\mathbf{B} / (n-1)$. Being symmetric and positive semi-definite, each element $C_{ij}$ of matrix $\mathbf{C}$ quantifies the correlation between the $i$-th and $j$-th samples within the data. The \textit{principal components} of the input data matrix $\mathbf{X}$ are the eigenvectors of $\mathbf{C}$, which define a change of coordinates in which the covariance matrix becomes diagonal:
\begin{equation}
    \mathbf{C}\mathbf{V} = \mathbf{V}\mathbf{\Lambda}
\end{equation}

The columns of the eigenvector matrix $\mathbf{V}$ are the principal components, and the elements of the diagonal matrix $\mathbf{\Lambda}$ represent the variances of the data along these directions, defining the most relevant components forming the new coordinate system. Based on this formulation, we define the autoencoder reconstruction with PCA pre-processing as
\begin{equation}
    \mathbf{\tilde{v}} = D \circ E(\mathbf{v})
\end{equation}
where $\mathbf{\tilde{v}}_i$ denotes the reconstruction of the $i$-th principal component in $\mathbf{V}$. The reconstructed high-dimensional data $\mathbf{\tilde{X}}$ is derived as
\begin{equation}
    \mathbf{\tilde{B}} = (\mathbf{B}\mathbf{V})\mathbf{\tilde{V}^*} \rightarrow     \mathbf{\tilde{X}} = [(\mathbf{X}-\overline{\mathbf{X}})\mathbf{V}]\mathbf{\tilde{V}^*} +\overline{\mathbf{X}}
\end{equation}

By applying the PCA pre-processing stage, the dimension of the data used to train the autoencoder model is reduced from ${\mathbb R}^{p\times n}$ to ${\mathbb R}^{n\times n}$, which represents a reduction of approximately two orders of magnitude in the input and output dimensions of the autoencoder network. Reducing the dimensionality of the data without losing information, as we keep all the SVD modes, decreases the sparsity of the input to the neural network, thereby reducing the number of training parameters and improving its efficiency. Hence, this paper examines the incorporation of the PCA pre-processing technique as described and compares its performance to the same model without its application.

\subsection{Surrogate modelling} \label{ss:surrogate}
The proposed surrogate model integrates the previously defined ML techniques to predict pressure fields from flight conditions. A $\beta-$VAE is employed to extract intrinsic features from the data within a reduced space, in which the GPR performs the regression. As discussed in \autoref{ss:PCA}, PCA data pre-processing is proposed to simplify the NN training, comparing the surrogate model's performance with and without PCA pre-processing.

\begin{figure*}[tb]
    \centering
    \includegraphics[width=0.9\linewidth]{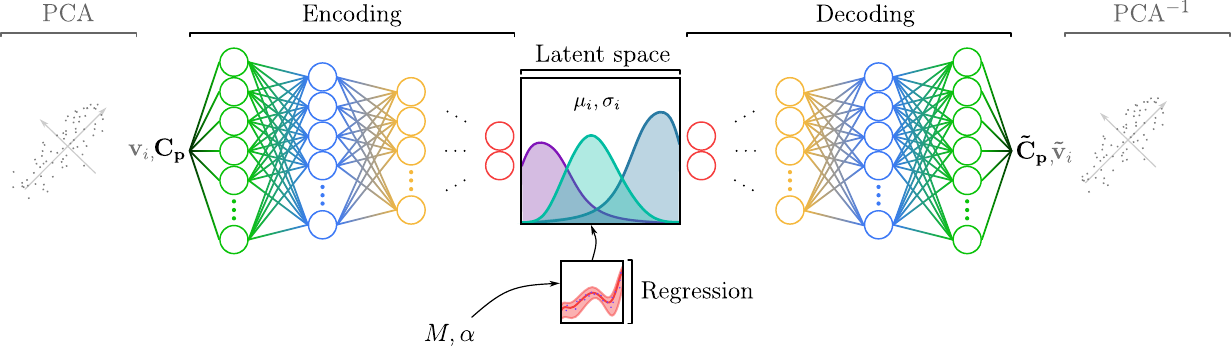}
    \caption{General outline of the model. The PCA obtains the principal components of the $C_p$ input matrix $\mathbf{X}$, $\mathbf{v}_i$, which are encoded into the latent space. The pressure fields are computed through the reconstruction of the decoder and the inverse PCA transformation. If PCA is not applied, the initial and final steps of the schematic do not apply.}
    \label{fig:scheme}
\end{figure*}

\autoref{fig:scheme} presents a high-level schematic of the model. For the PCA pre-processing case, the principal components $\mathbf{v}_i$ of the input data $\mathbf{X}$ are computed and used as inputs for the $\beta$-VAE. Conversely, the raw input data samples $\mathbf{x}$ are used directly when no pre-processing is considered. The encoder and decoder are trained simultaneously to minimize the loss function described in \autoref{ss:bVAE}. The encoder transforms the input data into a low-dimensional latent space, where each coordinate follows a Gaussian distribution $\mathbf{Z} \sim \mathcal{N}(\mu,\sigma)$. Once trained, the decoder reconstructs the input data from its latent representation, using only the mean value $\mu_i$. Once the $\beta$-VAE is trained and converged, NN parameters are fixed to train the regression model.

The architecture of the proposed $\beta$-VAE is summarized in \autoref{tab:model_architecture}. Because mesh connectivity is unavailable in the database under study, a multilayer perceptron (MLP) is chosen for both the encoder and decoder NNs with a symmetric hidden layer layout. To ensure a fair comparison between the $\beta$-VAE and the PCA$+\beta$-VAE strategies, the NNs' architectures are similarly designed, following a power-of-two pattern for gradually decreasing and then increasing the number of neurons, except for the input layer, which depends on the input data size. The activation function type, number of hidden layers, optimizer, and input data standardization are identical in both cases.  The main difference lies in the total number of parameters to be trained, which is 420 times higher without PCA pre-processing.

\begin{table}[tb]
    \centering
    \caption{$\beta-$VAE characteristics.}
    \label{tab:model_architecture}
    \renewcommand{\arraystretch}{1.5}
    \begin{tabular}{>{\raggedright\arraybackslash}p{0.44\linewidth-2\tabcolsep}
                >{\centering}p{0.21\linewidth-2\tabcolsep}
                >{\centering\arraybackslash}p{0.27\linewidth-2\tabcolsep}}
    \hline
    \textbf{Model}             & $\beta$-VAE                     & PCA+$\beta$-VAE \\
    \textbf{Optimizer}         & \multicolumn{2}{c}{ Adam }                        \\
    \textbf{Layers type}        & \multicolumn{2}{c}{ Fully connected }     \\
    \textbf{Activation type}   & \multicolumn{2}{c}{ ELU }                         \\
    \textbf{Hidden layers}     & \multicolumn{2}{c}{ 10 }                          \\[5pt]
    \textbf{\,\,\,\,\, Encoder}  & \multicolumn{1}{c}{ \makecell{1024-512-\\256-128-64} } & \multicolumn{1}{c}{ \makecell{256-128-\\64-32-16} }        \\[8pt]
    \textbf{\,\,\,\,\, Decoder} & \multicolumn{1}{c}{ \makecell{64-128-256-\\512-1024} } & \multicolumn{1}{c}{ \makecell{16-32-64-\\128-256} }        \\[5pt]
    \textbf{Input size}        & \multicolumn{1}{c}{49,574}      & 305             \\
    \textbf{No. of parameters} & \multicolumn{1}{c}{102,974,122} & 244,597         \\ \hline
    \end{tabular}
\end{table}

For the regression operation, we employ GPR within the latent space. To simplify the regression problem, GPR targets the means in the latent space, disregarding the standard deviation of each latent variable so that $z_i \simeq \mu_i$. GPR is trained to map the parameter space $\mathcal{P}$ to the latent space ($f: \mathcal{P} \rightarrow \mathcal{Z}$). Thus, the regression model predicts $z_i=\mu_i$ given the flight condition $[M,\alpha]_i$, such that $\hat{\mu}_i = f([M,\alpha]_i)$.

The kernel selection for GPR is based on the geometry of the latent space and its relationship with the parameter space. A linear kernel is proposed, supported by the observed linear solid relationship between the parameters and latent variables. This linear kernel effectively models direct relationships, providing a solid basis for regression tasks. Additionally, the linear kernel is combined with the Mat\'ern 3/2 kernel to capture the data set's nonlinearities. The Mat\'ern 3/2 kernel is well-suited for modeling smooth functions with significant nonlinear variations, making it appropriate for capturing the inherent complexities and variabilities in the data. Hence, the kernel is given by: 
\begin{equation} \label{eq:kernel}
k(\mathbf{p}, \mathbf{p'}) = k_{\text{linear}}(\mathbf{p}, \mathbf{p'}) \cdot k_{\text{Mat\'ern32}}(\mathbf{p}, \mathbf{p'}),
\end{equation}
being
\begin{align}
k_{\text{linear}}(\mathbf{p}, \mathbf{p'}) &= \sigma_{\text{linear}}^2 \mathbf{p}^{\mathsf{T}} \mathbf{p'},\\
k_{\text{Mat\'ern32}}(\mathbf{p}, \mathbf{p'}) &= \sigma_{\text{Mat\'ern32}}^2 \left(1 + \sqrt{3} \rho \right) \exp(-\sqrt{3} \rho),   
\end{align}
with  
\[
\rho = \frac{\|\mathbf{p} - \mathbf{p'}\|_2}{\ell_{\text{Mat\'ern32}}},
\]
where $\sigma_{\text{linear}}^2$, $\sigma_{\text{Mat\'ern32}}^2$, and $\ell_{\text{Mat\'ern32}}$ are the hyperparameters of the kernel.

Combining these kernels multiplicatively, the GPR model captures both linear and nonlinear dependencies, enhancing its predictive capability and ensuring a more accurate and robust representation of the relationships between flight conditions and latent space variables.

\section{Latent space analysis} \label{s:latent}
Despite the proven efficacy of $\beta$-VAEs in reducing the dimensionality of data into a latent space, optimizing the model involves several challenging aspects. These include determining the optimal architecture for the encoder and decoder, selecting the appropriate value for the hyperparameter $\beta$, and deciding the target dimension of the latent space $d$. With the architecture of the autoencoder fixed (see \autoref{ss:bVAE}), we conducted a parameter sweep of $d$ and $\beta$ to evaluate the model's performance and the interoperability of the latent space. Given that our database depends solely on flight conditions, represented by a tuple of two design variables ($[M,\alpha]$), it is reasonable to set the latent space dimension to $d=2$, which simplifies the regression from parameter space to latent space, $f: \mathcal{P}\in\mathbb{R}^2 \rightarrow \mathcal{Z}\in\mathbb{R}^2$. \citet{kang2022physics} proposed a \textit{physics-aware} $\beta$-VAE for transonic flow data over an airfoil, where the model is trained with a specific latent space dimension, followed by an assessment of the latent variables for physical relevance. They utilized a $\beta$-VAE with $d = 16$, finding that only two latent variables held significant physical meaning and highly correlated with the two design variables driving the problem: $M$ and $\alpha$. Considering this physical insight, we explored a $d\in[2,6]$ range. Regarding the regularization parameter $\beta$, we observed that small values are necessary to maintain an acceptable reconstruction loss. Consequently, we defined the range $\beta \in [0,10^{i}]$ with $i = -7, \ldots, 0$, allowing to assess the influence of $\beta$ by comparing it with the critical cases of $\beta=0$ and $\beta=1$. The former represents the limit of the $\beta$-VAE prioritizing accurate data reconstruction without the regularization and generative properties provided by the KL divergence term. At the same time, the latter corresponds to a classic VAE.

\begin{figure}[tb]
    \centering
    \includegraphics[width=0.99\linewidth]{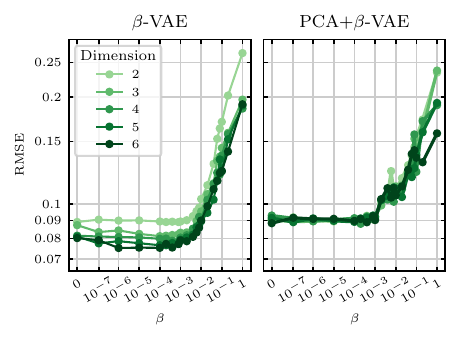}
    \caption{Reconstruction loss for $\beta$-VAE and PCA+$\beta$-VAE models depending on latent space dimension and $\beta$ hyperparameter for the testing dataset. Root mean square error is computed between original $C_p$ and reconstructed $\tilde{C_p}$ through the autoencoder.}
    \label{fig:choose_autoencoder}
\end{figure}

Figure \ref{fig:choose_autoencoder} shows the results of the parameter sweep for both $\beta$-VAE and PCA+$\beta$-VAE models, illustrating the root mean square error (RMSE) as a function of $\beta$ for different latent space dimensions. For both $\beta$-VAE and PCA+$\beta$-VAE models, the RMSE increases as the value of $\beta$ increases beyond a certain point. Lower values of $\beta$ generally result in lower RMSE, indicating better reconstruction accuracy. Comparing $\beta$-VAE and PCA+$\beta$-VAE, we observe that the PCA pre-processing yields a slightly higher RMSE regardless of the value of $\beta$. Nonetheless, the PCA pre-processing helps stabilize the model's performance by providing a more structured and compact input to the autoencoder and resulting in a simplified training of the model, which results in a collapsed value of RMSE independent of the latent space dimensions as $\beta$ is reduced.

The impact of the latent space dimension is evident in both cases. In the $\beta$-VAE model, the RMSE remains relatively stable for small values of $\beta$ across all dimensions but increases significantly for larger values of $\beta$. For the PCA+$\beta$-VAE model, the RMSE is more consistent across different dimensions, with a notable increase in RMSE for higher values of $\beta$. This effect implies that reducing the dimensionality of the data before $\beta$-VAE training minimizes the effect of $d$ on the reconstruction error. This result could stem from the orthogonality and hierarchical decomposition of the input data after PCA pre-processing. The principal components, being orthonormal and ordered by the amount of variance they capture, enable the autoencoder to focus on the two most relevant latent variables strongly correlated with the flight condition, as discussed later.

\begin{figure*}[tb]
    \centering
    \includegraphics[width = 0.90\linewidth]{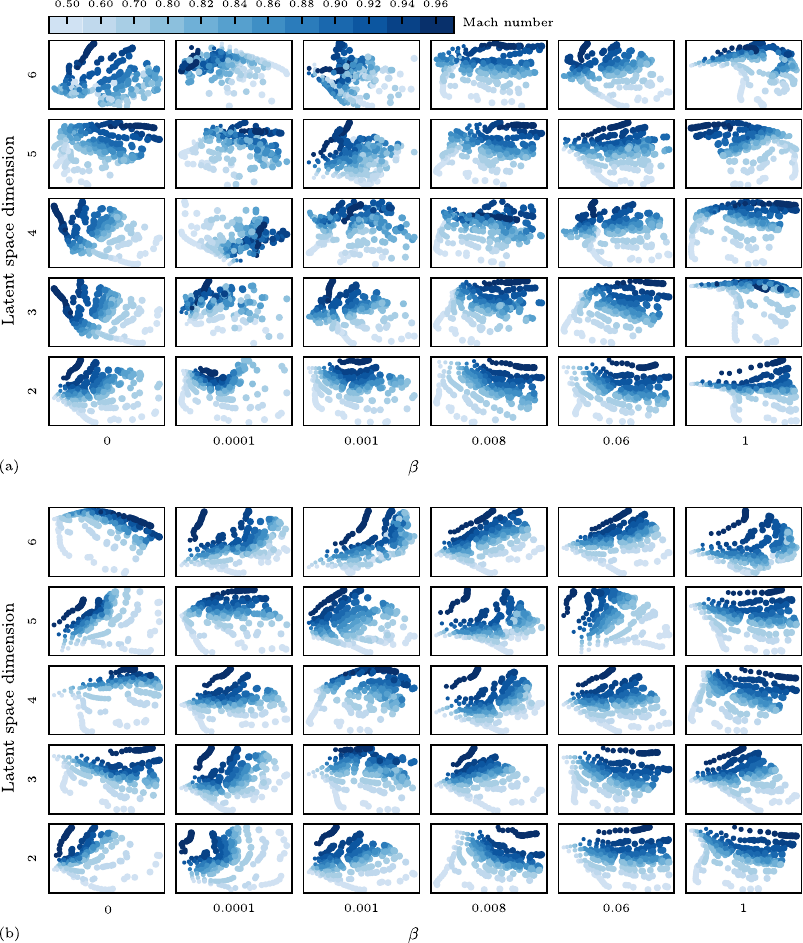}
    \caption{First two variables of the latent space depending on its dimension and $\beta$ hyperparameter for both (a) $\beta$-VAE and (b) PCA+$\beta$-VAE models. Value ranges are adjusted to be within the interval from -1 to 1. Latent spaces with more than two dimensions show the relationship between the first two modes that accumulate more information. Flight conditions are characterized by Mach number (point color) and $\alpha$ (point size).}
    \label{fig:latent_spaces_grid}
\end{figure*}

To further assess autoencoder performance, the physical interpretability of the latent space is crucial. The latent variables are ordered based on their contribution to the global reconstruction loss of the autoencoder, allowing the identification of the most relevant ones. To rank the latent variables, the model is evaluated by setting one mode to zero at a time. Modes that result in the highest error when set to zero are considered to contain the most information and are subsequently ordered  \cite{eivazi2022vae,solera2024beta}. Hence, the physical features contained in each LV can be assessed. \autoref{fig:latent_spaces_grid} shows the distribution of data in the latent space for both $\beta$-VAE (a) and PCA+$\beta$-VAE (b) models, focusing on the plane formed by the two most relevant latent variables. The color gradient represents the Mach number, while the size of each data point indicates $\alpha$. This visualization provides insights into how the latent variables capture the underlying flight conditions. For both $\beta$-VAE and PCA+$\beta$-VAE models, there is a perceptible distribution of data points that correlates with the flight conditions $[M,\alpha]$, which are well-represented in the latent space, indicating that the models effectively capture these critical aerodynamic parameters. For instance, in the case of $\beta=0.008$ for the $\beta$-VAE model, there is a clear correlation between the flight condition parameters and the latent variables $\mu_1$ and $\mu_2$: Mach number increasing from bottom to top and $\alpha$ from left to right. This result is noteworthy because the fully unsupervised autoencoder (no labels of the flight conditions are provided) learns to differentiate among flight conditions based on their associated pressure fields, leading to a smooth latent space ideal for the subsequent regression task. This behavior is consistent across different latent space dimensions and $\beta$ values. The PCA+$\beta$-VAE model shows a slightly more structured and stable distribution across different $\beta$ values, suggesting that the PCA pre-processing stage enables a better converged $\beta$-VAE training.

The reconstruction error in \autoref{fig:choose_autoencoder} is directly linked to the data distribution in the latent space (\autoref{fig:latent_spaces_grid}), strongly influenced by $\beta$. For minimal $\beta$ values (close to 0), both models show a well-organized latent space with distinct clustering according to flight conditions. As $\beta$ increases, the clustering becomes less distinct, particularly at $\beta=1$, where the regularization effect of the KL divergence term dominates, leading to a less informative latent space. The nonlinear relationship between $M$ and $\alpha$ is well captured at low $\beta$, but as $\beta$ increases and latent space orthogonality is promoted, the relationship between the parameter space $[M,\alpha]$ and latent variables $[\mu_1, \mu_2]$ becomes unclear. On the contrary, as the latent space dimension increases (from 2 to 6), the data distribution in the latent space remains relatively consistent, particularly for smaller values of $\beta$, which suggests that the most relevant features are adequately captured within the first two latent variables, supporting the decision to focus on this lower-dimensional representation. The remaining dimensions are energetically poor, with no significant information content.  This fact is especially evident when PCA pre-processing is introduced, as most latent spaces become visually sorted compared to the $\beta$-VAE model, illustrating the dimensionality reduction effect discussed in \autoref{fig:choose_autoencoder}. 

\begin{figure*}[tb]
    \centering
    \includegraphics[width=0.99\linewidth]{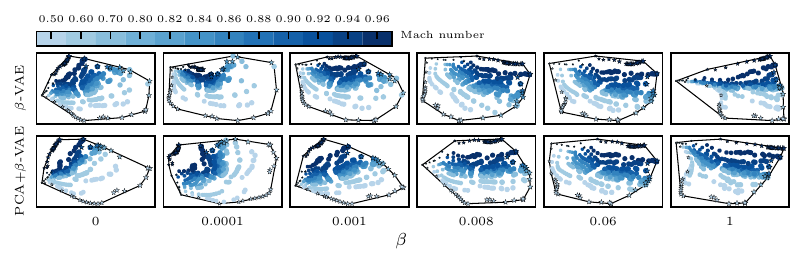}
    \caption{Latent spaces for $d=2$. Black lines \lcap{-}{black} highlight the boundaries of the latent space representation of the data. Star markers ($\star$) show boundary data based on flight conditions. Flight conditions are characterized by Mach number (point color) and $\alpha$ (point size).}
    \label{fig:convex_hull}
\end{figure*}

The choice of latent space dimension $d$ does not significantly affect the RMSE for lower values of $\beta$. While increasing the dimensionality can capture more information, it requires careful tuning of $\beta$ to avoid over-regularization. Choosing a small $\beta$ value and using PCA pre-processing can yield better surrogate models with higher accuracy and stability for practical applications. Based on this analysis, the latent space dimension is fixed at $d=2$ for the remainder of the study, establishing a direct relationship between the dimensionality of the parameter space and the latent space. Such a simplified representation of the data in a two-dimensional latent space facilitates the interpretability of the results and the regression operation in the surrogate.

On the other hand, selecting an optimal $\beta$ is complex, without any physical guidelines, and based on the RMSE only.  The range $\beta \in [0, 10^{-3}]$ results in a plateau with the lowest RMSE for both models, with a reasonable data distribution in the latent space for regression. Thus, small values of $\beta$ are crucial for maintaining reconstruction accuracy. As $\beta$ approaches 1, the RMSE increases sharply, reflecting the trade-off between reconstruction accuracy and the regularization imposed by the KL divergence term. These observations suggest that both models perform best with small values of $\beta$, confirming the need to minimize the regularization term to maintain reconstruction fidelity. Nonetheless, the remaining analysis in this manuscript will consider several $\beta$ options to show the performance of the surrogate model based on this parameter.

To conclude the latent space analysis, and with the latent space dimension fixed at $d=2$, it is insightful to visualize the boundary data of the parameter space to confirm how well the physical parameter domain is represented in the latent space. \autoref{fig:convex_hull} illustrates the distribution of data in the latent space for various values of $\beta$ and for both models, $\beta$-VAE (top row) and PCA+$\beta$-VAE (bottom row). The marker size and color denote $\alpha$ and $M$, respectively, while star markers ($\star$) represent the boundary data from the flight envelope (see \autoref{fig:dataset_wing}a). Such boundary data points, which represent the extreme flight conditions, lie on or within the boundaries of the latent space for both models (black lines). This analysis indicates that the latent spaces generated by both models adequately confine the entire range of flight conditions. 

Regardless of the value of $\beta$ and the PCA pre-processing, all the star marks are near the actual boundary of the latent space. A few specific points fall on the boundary despite not being associated with the extreme conditions of the Mach and $\alpha$ envelope.  This representation also reveals that the boundary data is not uniformly distributed across the latent space boundary. The latent space shrinks in regions corresponding to high Mach numbers, suggesting that the pressure distribution over the wing is more influenced by Mach number than by angle of attack. Nonetheless, since the data distribution in the latent space lies within the boundaries, no major issues will arise due to extrapolation when regression is performed.

\section{Evaluation of the surrogate model} \label{s:results}
The combination of PCA pre-processing (if considered), the $\beta$-VAE, and the GPR within the latent space concludes in a global surrogate model capable of predicting the whole pressure distribution over a wing given the flight condition in terms of $\alpha$ and Mach. This sophisticated yet efficient model is evaluated in this section. First, the regression task within the latent space is assessed, followed by the performance evaluation of the whole surrogate framework. Eventually, we propose a novel strategy to tune the surrogate model, reducing the dependency on $\beta$ while improving the overall predictive accuracy of the surrogate.

\begin{figure*}[tb]
    \centering
    \includegraphics[width=0.99\linewidth]{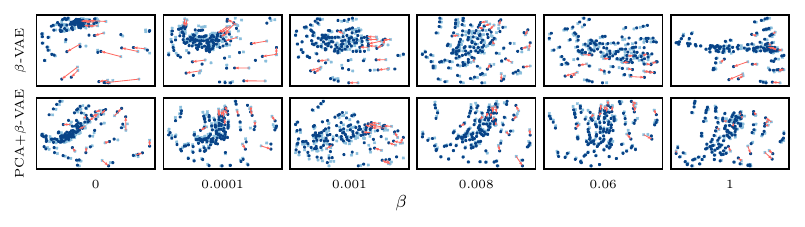}
    \caption{Regression performance in the latent space on the test set, comparing the reference latent space \sy{quiver_ref_color}{x} and the prediction \sy{quiver_pred_color}{o*}. The arrows ($\color{quiver_arrow_color} \rightarrow$) highlight the displacement of the predictions for the 15 cases with the highest error.}
    \label{fig:quiver}
\end{figure*}
\subsection{Regression within the latent space} \label{ss:regression}
The regression model must accurately predict the latent variables yielded by the encoder from their associated flight condition. The kernels used for the GPR model (defined in \autoref{ss:surrogate}) should capture the evident relation between the parameter space and the latent space, as outlined in \autoref{s:latent}. For the sake of clarity, we will refer to the latent space provided by the autoencoder and visualized in \autoref{fig:convex_hull} as \textit{reference} latent space, which is the target to predict by the GPR model. \autoref{fig:quiver} shows the regression performance in the latent space on the test set, comparing the reference latent space with the predicted latent space. The arrows indicate the displacement of the predicted latent variables for the 15 cases with the highest prediction error, visually representing the prediction error. The regressor performs well, with most predictions aligning closely with the reference latent space points, indicating that the regression model can effectively map the parameter space to the latent space for most cases.

The regressor tends to fail more frequently at points near or on the boundaries of the parameter space, especially at high angles of attack (the right side of the latent space). This pattern suggests that the regressor struggles more in these regions, possibly due to the higher complexity and nonlinearity of the aerodynamic responses, such as flow separation or displacement of shockwaves. Additionally, these boundary data points in the latent space have fewer neighbors, reducing the local information available for the regressor and making accurate predictions more challenging.

Comparing the $\beta$-VAE model (top row) with the PCA+$\beta$-VAE model (bottom row), it is evident that PCA pre-processing enhances the stability and accuracy of the regression model. The PCA+$\beta$-VAE model shows smaller arrows, indicating lower prediction errors across different $\beta$ values. PCA pre-processing helps the autoencoder to learn meaningful representations with a better distribution of samples in the latent space, aiding the regression model in learning a more accurate mapping function from the parameter space to the latent space.

For minimal values of $\beta$ (0 and 0.0001) in the $\beta$-VAE model, most test data concentrate in a single cluster, with some cases far from the rest due to the lack of regularization. These outlier cases, mainly defined by high angles of attack, lead to considerable displacements in the latent space predictions. For the PCA+$\beta$-VAE model and $\beta > 0.001$ in the $\beta$-VAE model, the effect of the hyperparameter $\beta$ is less critical since the latent space is already well-distributed, exhibiting a clear correlation with the flight condition parameters (as discussed in \autoref{s:latent}).

For a quantitative analysis, the performance of the regressor is evaluated in terms of the normalized root mean square error (NRMSE) between the reference latent space and the regression prediction for both training and testing datasets. This metric quantifies the prediction error relative to the overall size of the latent space, which is suitable for comparing the results of different latent spaces since their extension is not invariant with $\beta$. Given a data sample in the latent space $\mathbf{\mu}$, its mean value $\bar{\mathbf{\mu}}$, and its prediction $\hat{\mathbf{\mu}}$, the NRMSE is defined as:
\begin{equation} 
\text{NRMSE}(\mu,\hat{\mu}) = \dfrac{\norm{\mathbf{\mu} - \hat{\mathbf{\mu}}}_2}{\norm{\mathbf{\mu} - \bar{\mathbf{\mu}}}_2}. 
\end{equation}

\begin{figure}[tb]
    \centering
    \includegraphics[width=0.99\linewidth]{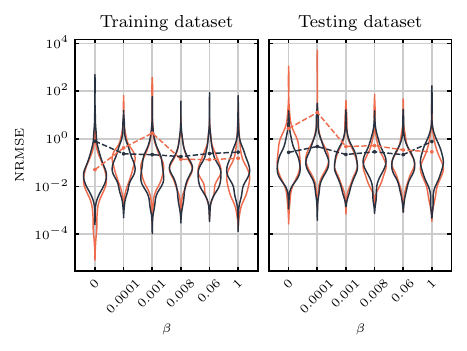}
    \caption{Normalized root mean squared error between the reference latent space and the regression prediction from flight conditions, for both $\beta$-VAE \lcap{_}{no_pca_color} and PCA+$\beta$-VAE \lcap{_}{pca_color} models. Dashed lines \lcap{--}{black} indicate mean values.}
    \label{fig:nrmse_regressor}
\end{figure}
\autoref{fig:nrmse_regressor} depicts the prediction NRMSE distribution as a function of $\beta$ for both $\beta$-VAE and PCA+$\beta$-VAE models, across training and testing datasets. The dashed lines represent the mean error evolution with $\beta$ value, while the violin plots show the error distribution for each $\beta$. As $\beta$ decreases, the error distribution widens, reflecting greater variability in prediction errors. These tails in the error distribution are due to outliers, consistent with the large displacements reported in \autoref{fig:quiver} for high-$\alpha$ cases. Nonetheless, the mean NRMSE remains fairly constant and independent of $\beta$.

The PCA+$\beta$-VAE model consistently shows lower mean NRMSE and narrower error distributions compared to the $\beta$-VAE model, reinforcing the benefit of PCA pre-processing. Additionally, although there is no evidence of overfitting, the prediction error in the testing set is almost equal to the training set for the PCA+$\beta$-VAE model, whereas it becomes larger without PCA pre-processing.

In conclusion, the performance of the GPR model and the selected kernels is reasonably accurate regardless of PCA pre-processing and the $\beta$ value, justifying its choice despite the existence of more sophisticated regressors, such as DNN-based regressors. The GPR robustness and versatility make it a widely employed choice in similar studies \cite{kang2022physics, NietoCentenero2023fusing}.

\subsection{Performance in predicting aerodynamic data} \label{ss:enclosure}
The overall performance of the surrogate model is evaluated by predicting the pressure distribution on the upper wing of the XRF1 aircraft for various combinations of Mach number and angle of attack within the flight envelope (see \autoref{fig:dataset_wing}a). The prediction process involves feeding the flight condition ($M, \alpha$) to the GPR model, which predicts the latent representation of the pressure distribution. Then, the predicted latent representation is mapped into the high-dimensional space for the previously trained decoder.  The inverse PCA is applied for the cases with PCA pre-processing to obtain the final $C_p$ distribution. For comparison, the surrogate models are benchmarked against a direct multilayer perceptron regressor that directly maps from the parameter space to the pressure distribution.

\begin{figure}[tb]
    \centering
    \includegraphics[width=0.99\linewidth]{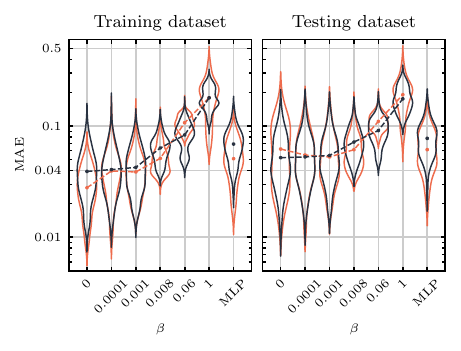}
    \caption{Mean absolute error between ground truth and $C_p$ predictions for $\beta$-VAE+GPR \lcap{_}{no_pca_color} and PCA+$\beta$-VAE+GPR\lcap{_}{pca_color} models. The last column corresponds to a direct MLP regressor (from flight conditions to coefficients of pressure), whose architecture mimics the decoder from the autoencoder models. Dashed lines \lcap{--}{black} indicate mean values.}
    \label{fig:full_workflow}
\end{figure}

\autoref{fig:full_workflow} displays the Mean Absolute Error (MAE) between the ground truth ($C_p$) and predictions ($\hat{C}_p$) for $\beta$-VAE+GPR and PCA+$\beta$-VAE+GPR models across different $\beta$ values for both training and testing datasets. The PCA+$\beta$-VAE model generally shows lower MAE than the $\beta$-VAE model, indicating that PCA pre-processing improves the accuracy of $C_p$ predictions. For both models, smaller $\beta$ values (e.g., 0, 0.0001) yield lower MAE, suggesting that minimal regularization results in better performance. As $\beta$ increases, the average MAE tends to increase despite decreasing variability, reflecting over-regularization's adverse impact.  The violin plots indicate that the variability in MAE is generally lower for the PCA+$\beta$-VAE model, demonstrating more consistent performance across different flight conditions. Comparing the performance of the surrogates against the MLP benchmark (depicted in the last column), the PCA+$\beta$-VAE model shows comparable or better performance than the MLP model for low $\beta$ values, highlighting its efficacy in capturing aerodynamic characteristics more accurately. Nonetheless, over-regularizing the latent space is detrimental to overall predictive performance, making it less advantageous to consider the additional sophistication of the autoencoder versus the direct regression with the MLP. It should be noted that the direct use of MLP for this type of application suffers from a more significant scalability limitation than the PCA+$\beta$-VAE model due to the curse of dimensionality caused by the size of the output data, and in our model is mitigated by the use of PCA pre-processing, which turns out to be less effective for the MLP model.

To further analyze the performance of the surrogate, \autoref{tab:performance_indicators} summarizes the overall performance of the surrogate models in terms of MAE, RMSE, and $R^2$ score using the testing dataset. A significant observation is the considerable difference in memory requirement between the surrogate models with and without PCA pre-processing. The dimensionality reduction achieved by PCA reduces the overall surrogate's size and the associated computational cost. This efficiency is mainly due to the smaller number of parameters needed to assemble the encoder and decoder networks, which is much smaller than using the high-dimensional $C_p$ data directly as input. Considering that a total of 210 models have been tested in this study, this size difference is crucial. Indeed, the model with PCA pre-processing is over four hundred times smaller in memory size, making it easy to train on any modern personal computer.

\begin{table}
    \centering
    \caption{Performance indicators of the models. They include MAE, RMSE, and $R^2$ scores, which have been computed using the testing dataset.}
    \label{tab:performance_indicators}
    \renewcommand{\arraystretch}{1.5} 
    \begin{tabular}{>{\centering}p{0.141\linewidth-2\tabcolsep}
                >{\centering}p{0.140\linewidth-2\tabcolsep}
                >{\centering}p{0.140\linewidth-2\tabcolsep}
                >{\centering}p{0.137\linewidth-2\tabcolsep}
                >{\centering}p{0.140\linewidth-2\tabcolsep}
                >{\centering}p{0.140\linewidth-2\tabcolsep}
                >{\centering\arraybackslash}p{0.137\linewidth-2\tabcolsep}}    
    \hline
    \textbf{Model} & \multicolumn{3}{c}{$\beta$-VAE} & \multicolumn{3}{c}{PCA+$\beta$-VAE} \\ 
    \hline
    \textbf{Size} & \multicolumn{3}{c}{1.15 GB} & \multicolumn{3}{c}{2.84 MB} \\ 
    \hline
    $\beta$ & \multicolumn{1}{c}{\textit{MAE}} & \multicolumn{1}{c}{\textit{RMSE}} & \multicolumn{1}{c}{$R^2$} & \multicolumn{1}{c}{\textit{MAE}} & \multicolumn{1}{c}{\textit{RMSE}} & \multicolumn{1}{c}{$R^2$}  \\ 
    \hline
    0      & 0.062 & 0.114 & 0.812 & 0.052 & 0.094 & 0.875 \\
    0.0001 & 0.055 & 0.101 & 0.854 & 0.053 & 0.095 & 0.870 \\
    0.001  & 0.053 & 0.098 & 0.864 & 0.054 & 0.096 & 0.869 \\
    0.008  & 0.061 & 0.102 & 0.852 & 0.072 & 0.111 & 0.821 \\
    0.06   & 0.110 & 0.155 & 0.671 & 0.091 & 0.138 & 0.735 \\
    1      & 0.192 & 0.266 & 0.106 & 0.176 & 0.234 & 0.363 \\
    \hline
    \end{tabular}
\end{table}

The performance metrics reveal interesting differences between the models. Generally, errors are lower when PCA pre-processing is applied, and the $R^2$ score is higher despite having fewer trainable parameters. This result confirms that PCA pre-processing enhances model accuracy both regarding overall performance and maximum errors within the $C_p$ distribution. The $R^2$ scores indicate that the PCA+$\beta$-VAE model explains a higher proportion of data variance than the $\beta$-VAE model. The improved performance due to PCA pre-processing can be attributed to the reduced dimensionality gap between the latent space and the high-dimensional $C_p$ space, allowing the model to retain the physical meaning of the latent space. This performance makes PCA+$\beta$-VAE a preferable option over $\beta$-VAE, considering its considerably lower computational effort.

Combining the insights from \autoref{fig:full_workflow} and \autoref{tab:performance_indicators}, it is evident that the PCA+$\beta$-VAE+GPR model outperforms the $\beta$-VAE+GPR model across all performance metrics. The benefits of PCA pre-processing are clear: \textbf{enhanced accuracy}, with lower MAE and RMSE values, coupled with higher $R^2$ scores; \textbf{consistency}, indicated by narrower error distributions in the violin plots; and \textbf{efficiency}, due to the significantly smaller model size. Moreover, comparing the proposed surrogates with the MLP benchmark concludes that introducing the autoencoder benefits the overall performance only if the value of $\beta$ is tuned correctly. Consistent with the discussions in \autoref{s:latent}, a lower $\beta$ is preferred regarding autoencoder performance (reconstruction capabilities) and overall surrogate performance.

\begin{figure*}[tb]
    \centering
    \includegraphics[width=0.95\linewidth]{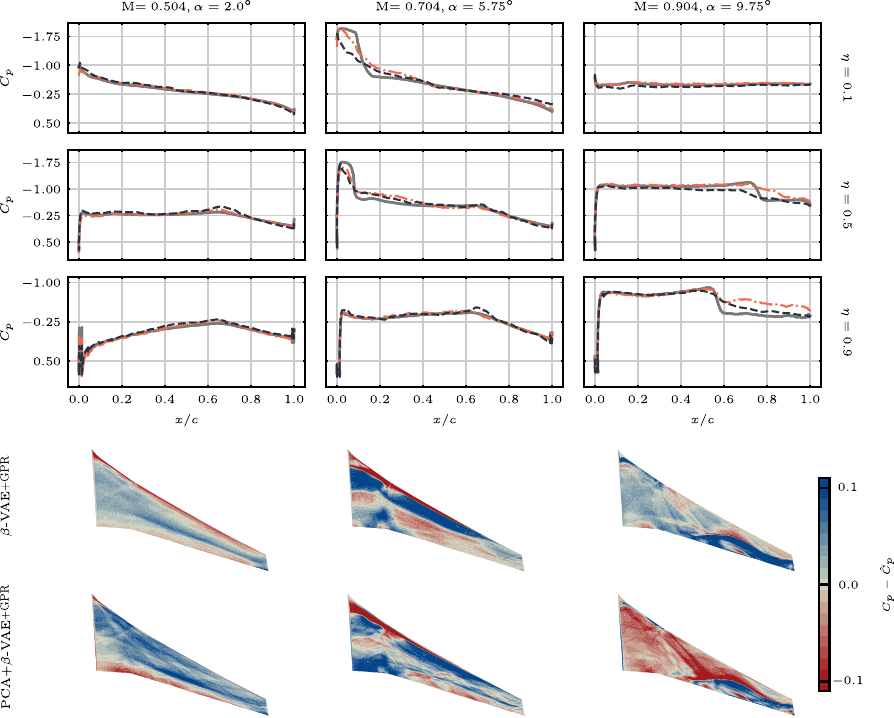}
    \caption{Difference between ground truth and predicted coefficients for the visualization test cases with $\beta = 0.008$. Chordwise pressure distributions from $\beta$-VAE+GPR \lcap{.-}{no_pca_color} and  PCA+$\beta$-VAE+GPR \lcap{--}{pca_color} models at span percentages $\eta =$ 0.1, 0.5, 0.9 are displayed and contrasted with ground truth \lcap{-}{gray_plane}.}
    \label{fig:wing_plot}
\end{figure*}

The global metrics of the models are contrasted with the predicted pressure distributions on the so-called visualization cases. For clarity, only the surrogate models for $\beta=0.008$ are considered, which seems to be a good trade-off in predicting performance, latent space reconstruction and interoperability. Thus, \autoref{fig:wing_plot} presents a detailed comparison of the ground truth pressure coefficient ($C_p$) fields obtained from RANS simulations and the predicted values ($\hat{C}_p$) from the surrogate models for three testing flight conditions. The figure includes contour maps of the prediction error ($C_p - \hat{C}_p$) over the wing and chordwise distributions at three spanwise locations ($\eta = 0.1$, $0.5$, $0.9$).

For the low ($M,\alpha$) case, both models generally exhibit small prediction errors across the wing surface, with some localized errors near the leading edge due to minor inaccuracies in capturing the leading-edge suction peak. Both models tend to slightly underestimate $C_p$ on the wing surface, although the leading and trailing edges show some overestimation.
The prediction errors increase slightly in the moderate condition case, particularly near the leading edge. Both models fail to capture the abrupt change in pressure at the suction peak, leading to a smoother than reference pressure distribution. The PCA+$\beta$-VAE model has slightly smaller error magnitudes on average, despite some regions, such as the leading edge near the wing root, showing higher errors. 
For the high Mach-high $\alpha$ case, prediction errors are higher, although very localized. Both models struggle with the sudden pressure drop at the shockwave region, though the PCA+$\beta$-VAE model generally exhibits lower errors than the $\beta$-VAE model.

Accurately capturing the leading edge suction peak is challenging due to high-pressure gradients, as indicated by localized errors near the leading edge. Predicting the location and strength of shockwaves is also critical, especially at higher Mach numbers and angles of attack. Overall, the performance of both surrogates is very similar in global metrics. This similarity was already evident in \autoref{fig:choose_autoencoder}, where the reconstruction errors of both autoencoder models were approximately the same. Given that the GPR model performs similarly in both cases, the resulting surrogates also perform similarly. The PCA+$\beta$-VAE model shows improved performance in critical regions, likely due to better latent space representation facilitated by PCA pre-processing. Moreover, it is worth highlighting the computational efficiency achieved with the PCA pre-processing and the greater interpretability of the latent space, especially at low $\beta$ values, where the surrogates perform best.

The proposed surrogate framework combining $\beta$-VAE with GPR outperforms classic surrogate models based on dimensionality reduction techniques. In particular, this dataset has been also tested for Isomap+DNN and POD+DNN models in \citet{castellanos2022assessment}, confirming the enhanced accuracy achieved by $\beta$-VAE despite the fact of using a classic regressor as it is GPR.

\subsection{Fine-tuning process: towards $\beta$-independent surrogate model} \label{ss:finetunning}
A fine-tuning process is introduced to enhance the surrogate model's accuracy. This process involves retraining the previously fixed decoder using the latent spaces predicted by the GPR model, allowing the decoder to learn and correct regression errors. Starting from the previously converged decoder, which effectively reconstructed $C_p$ from its latent representation (or principal components in the case of PCA pre-processing), the decoder's parameters are fine-tuned. The critical difference in this retraining phase is that the decoder now uses the predicted latent representation $\hat{\mu}$ from GPR instead of the actual latent representation $\mu$. Consequently, the decoding process is no longer strictly tied to the latent space distribution dictated by the KL divergence term of \autoref{ec:betaVAEloss} but to its approximation predicted by the GPR model from the parameter space.

\begin{figure}[tb]
    \centering
    \includegraphics[width=0.99\linewidth]{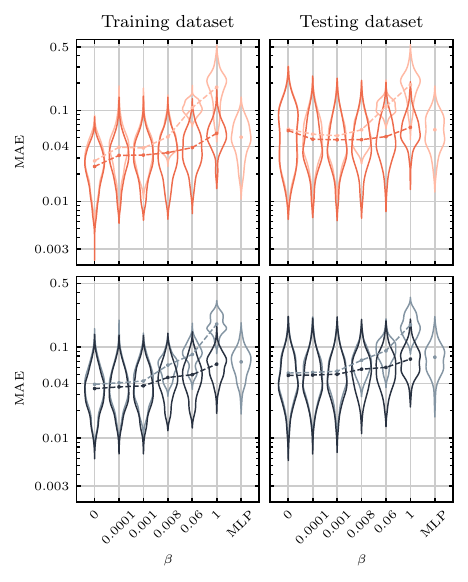}
    \caption{Mean absolute error between ground truth and $C_p$ predictions for fine-tuned $\beta$-VAE+GPR \lcap{_}{no_pca_color} and PCA+$\beta$-VAE+GPR\lcap{_}{pca_color} models. The last column corresponds to a direct MLP benchmark. Lighter colors indicate original models (see \autoref{fig:full_workflow}) for reference. Dashed lines \lcap{--}{black} indicate mean values.}
    \label{fig:full_workflow_ft}
\end{figure}

The fine-tuned $\beta$-VAE+GPR and PCA+$\beta$-VAE+GPR models are compared with the original in \autoref{fig:full_workflow_ft}. Post-fine-tuning, the error distributions shift lower, achieving better metrics than the direct-MLP benchmark model, regardless of $\beta$. The improvement of both surrogates over the benchmark justifies the development of a sophisticated ensemble model, which is more efficient in terms of time and computational requirements.

Although the violin plots show a wider distribution of MAE for each $\beta$, the average performance of the surrogate model is enhanced by the tuned decoder for both models. It is worth noting that there is no appreciable overfitting of the training data since the increased performance is also achieved for the testing set. The improvement is especially noticeable for large values of $\beta$ as discussed in \autoref{s:latent}, high $\beta$ reduces the reconstruction performance of the autoencoder despite regularizing the latent space. The fine-tuning process mitigates this by allowing the new decoder to handle the reconstruction error. As a result, the fine-tuning process reduces the model's dependency on the $\beta$ parameter, keeping MAE bounded between 0.05 and 0.07. This result suggests that the fine-tuned decoder can generalize better without extensive $\beta$ sweeps, acting as a pseudo-decoder, less tied to the KL divergence.

\autoref{tab:performance_indicators_ft} shows the performance indicators (MAE, RMSE, and $R^2$) of the fine-tuned models, which can be directly compared to \autoref{tab:performance_indicators}. There is a significant improvement in metrics, especially at higher $\beta$ values, with MAE reductions ranging from $\sim$3\% to $\sim$66\%. At first glance, the fine-tuning has a more significant  impact on the surrogate without PCA pre-processing. Since in the $\beta$-VAE+GPR model, the room for improvement is considerably more significant; the fine-tuning process is not only coping with the regression error from the GPR but with the reconstruction errors reported in \autoref{fig:choose_autoencoder}.  Moreover, the encoder's NN without PCA pre-processing has a considerably larger number of parameters; hence, fine-tuning the NN yields better performance.Nonetheless, the overall performance of both models is very similar, recalling the reduction in memory by a factor of 420 when PCA is applied.

\begin{table}[bt]
    \centering
    \caption{Performance indicators of the fine-tuned models. They include MAE, RMSE, and R$^2$ scores, which have been computed using the testing dataset.}
    \label{tab:performance_indicators_ft}
    \renewcommand{\arraystretch}{1.5} 
    \begin{tabular}{>{\centering}p{0.141\linewidth-2\tabcolsep}
                >{\centering}p{0.140\linewidth-2\tabcolsep}
                >{\centering}p{0.140\linewidth-2\tabcolsep}
                >{\centering}p{0.137\linewidth-2\tabcolsep}
                >{\centering}p{0.140\linewidth-2\tabcolsep}
                >{\centering}p{0.140\linewidth-2\tabcolsep}
                >{\centering\arraybackslash}p{0.137\linewidth-2\tabcolsep}}   
    \hline
    \textbf{Model} & \multicolumn{3}{c}{$\beta$-VAE} & \multicolumn{3}{c}{PCA+$\beta$-VAE} \\ 
    \hline
    $\beta$ & \multicolumn{1}{c}{\textit{MAE}} & \multicolumn{1}{c}{\textit{RMSE}} & \multicolumn{1}{c}{$R^2$} & \multicolumn{1}{c}{\textit{MAE}} & \multicolumn{1}{c}{\textit{RMSE}} & \multicolumn{1}{c}{$R^2$}  \\ 
    \hline
    0      & 0.060 & 0.112 & 0.818 & 0.049 & 0.091 & 0.883 \\
    0.0001 & 0.048 & 0.093 & 0.876 & 0.049 & 0.092 & 0.881 \\
    0.001  & 0.048 & 0.092 & 0.881 & 0.050 & 0.092 & 0.879 \\
    0.008  & 0.048 & 0.090 & 0.884 & 0.057 & 0.098 & 0.860 \\
    0.06   & 0.052 & 0.093 & 0.876 & 0.060 & 0.102 & 0.850 \\
    1      & 0.065 & 0.107 & 0.832 & 0.074 & 0.117 & 0.796 \\
    \hline
    \end{tabular}
\end{table}

Focusing on the performance indicators in \autoref{tab:performance_indicators_ft}, the fine-tuned decoder leads to better performance regardless of $\beta$. The $R^2$ scores indicate that the tuned models explain a higher proportion of variance in the data. Accuracy is also enhanced, with lower averaged MAE and RMSE values. However, the variability of the error values (violin plots) becomes wider for larger improvements through fine-tuning, particularly at high $\beta$.

The enhancement gained through fine-tuning is intrinsically linked to the nature of the encoding process. For small $\beta$ values, the KL restriction is neglected during training (\autoref{ec:betaVAEloss}), leading to latent space coordinates with a standard deviation $\sigma$ close to zero. 
Conversely, at high values of $\beta$, the model is over-regularized, leading to a loss of accuracy in the reconstruction due to its lower relative weight in the loss function. These models, therefore, offer a greater margin for improvement during the fine-tuning process, as the regularization is now removed, and the loss function only includes the reconstruction term.

\begin{figure}[tb]
    \centering
    \includegraphics[width=0.95\linewidth]{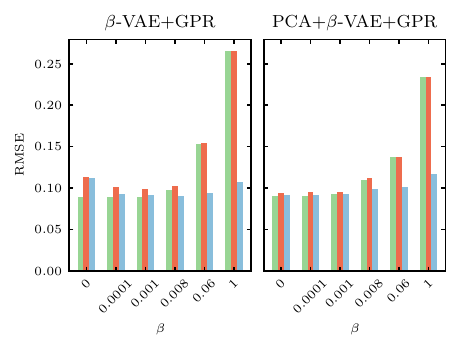}
    \caption{RMSE comparison between autoencoder \lcap{_}{ae_error_color}, regressor + decoder \lcap{_}{full_model_error_color}, and fine-tuned regressor + decoder \lcap{_}{full_ft_model_error_color} reconstructions for both $\beta$-VAE+GPR and PCA+$\beta$-VAE+GPR models.}
    \label{fig:rmse_barplot}
\end{figure}

\begin{figure*}[tb]
    \centering
    \includegraphics[width=0.99\linewidth]{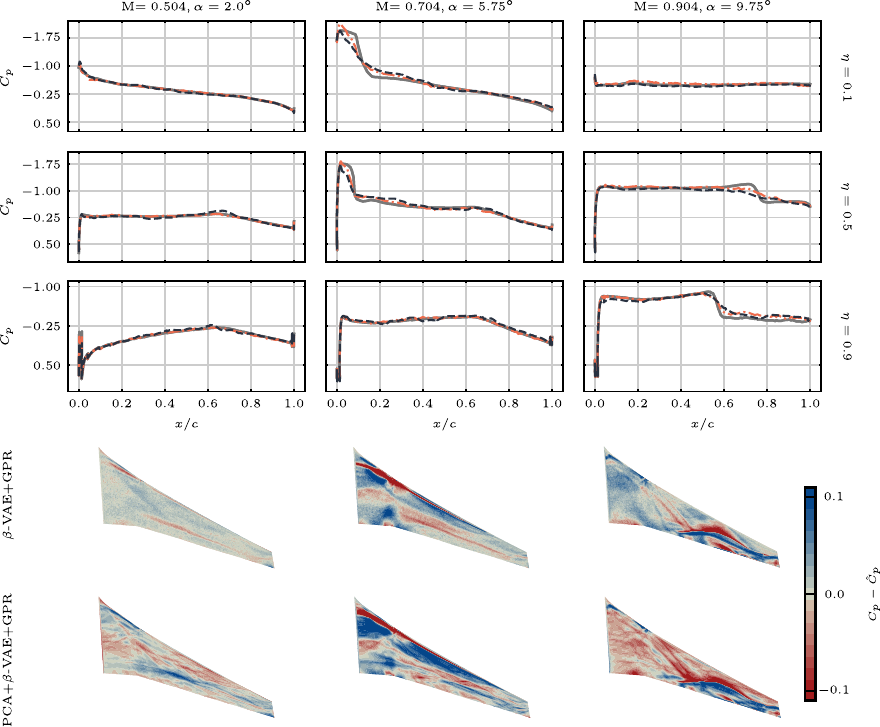}
    \caption{Difference between ground truth and predicted coefficients for the visualization test cases with $\beta = 0.008$. Chordwise pressure distributions from fine-tuned $\beta$-VAE+GPR \lcap{.-}{no_pca_color} and  PCA+$\beta$-VAE+GPR \lcap{--}{pca_color} models at span percentages $\eta =$ 0.1, 0.5, 0.9 are displayed and contrasted with ground truth \lcap{-}{gray_plane}.}
    \label{fig:wing_plot_ft}
\end{figure*}

The fine-tuning process is tailored to each $\beta$ case, as illustrated in \autoref{fig:rmse_barplot}. For small $\beta$ values, the autoencoder is already well-trained in reconstruction (\autoref{fig:choose_autoencoder}), so fine-tuning primarily addresses the regression error. Given that the regression error (\autoref{fig:nrmse_regressor}) is reasonably low and nearly invariant with $\beta$, the error before and after fine-tuning remains similar and higher than the autoencoder's reconstruction error alone. Consequently, for low $\beta$, the fine-tuned decoder mainly compensates for the regression error, resulting in limited improvement.

On the other hand, for larger $\beta$ values, the reconstruction loss is the predominant contributor to the global surrogate error, which is mitigated by retraining. As $\beta$ increases, the KL divergence term becomes more influential, and the fine-tuning process significantly enhances the reconstruction capacity of the decoder. Consequently, it results in better overall metrics than the autoencoder's standalone performance, as the fine-tuned decoder adapts to the generalized values of the latent space.

Similar to \autoref{fig:wing_plot}, the performance of the fine-tuned surrogate models in predicting pressure distributions ($C_p$) on the wing is assessed for the visualization cases in \autoref{fig:wing_plot_ft}. Fine-tuning reduces prediction errors across the wing surface for both models. Localized errors near the leading edge are minimized, indicating a better capture of the leading-edge suction peak. The overall prediction of $C_p$ is improved, with reduced underestimation and overestimation at the leading and trailing edges. In the moderate ($M,\alpha$) case, the pressure distribution closely matches the ground truth, with the PCA+$\beta$-VAE model continuing to show improved performance and reduced error magnitudes. Additionally, for the high ($M,\alpha$) case, fine-tuning enhances the model's ability to predict the pressure drop at the shockwave region. Both models exhibit lower errors, with the PCA+$\beta$-VAE+GPR model showing the most significant improvement, though there is still a tendency to smooth the sudden pressure drop.

Fine-tuning the decoder enhances the surrogate model's performance in predicting pressure distributions on the wing across all flight conditions, particularly for higher Mach numbers and angles of attack where aerodynamic complexity increases. This process reduces localized errors in critical regions such as the leading edge and shockwave areas, providing more accurate $C_p$ predictions that closely match the ground truth.

In summary, the fine-tuning process significantly improves the surrogate models' ability to predict complex aerodynamic pressure distributions, making them more reliable and efficient for practical applications. This novel approach paves the way for a $\beta$-independent surrogate model using $\beta$-VAEs. It allows the selection of $\beta$ to enhance the interpretability of the latent space, facilitating the extraction of physical insights or performing complex tasks like multifidelity analysis while ensuring excellent prediction performance. Therefore, users can choose the $\beta$ value based on additional requirements rather than prediction accuracy, as the fine-tuning strategy mitigates potential errors.

\section{Conclusions} \label{s:conclusions}
An analysis of the application of $\beta$-variational autoencoders combined with regression based on Gaussian processes has been conducted to predict pressure fields over a wing subjected to transonic flight conditions. The complexity of this three-dimensional flow allows us to evaluate the capability of data-driven models to predict nonlinearities in pressure fields due to shockwaves, high angles of attack, and the effects of high load factor conditions.

A multilayer perceptron architecture is proposed for the encoder and decoder neural networks of the $\beta$-VAE with a symmetric layout, taking high-dimensional $C_p$ samples as input and providing their latent representation at the autoencoder's bottleneck. A comprehensive analysis of the latent space was performed, exploring different latent space dimensions and $\beta$ values. Additionally, PCA pre-processing was introduced to enable the $\beta$-VAE model to focus on reconstructing the principal components of $C_p$, reducing dimensionality and the complexity of the neural network.

The latent space analysis demonstrates that both $\beta$-VAE and PCA+$\beta$-VAE models effectively capture key aerodynamic features, strongly correlating with flight conditions.  PCA pre-processing stabilizes model performance by providing a structured and compact input to the autoencoder. This methodology simplifies training and results in consistent reconstruction errors independent of latent space dimensions as $\beta$ is reduced. However, over-regularizing the latent space (high $\beta$) diminishes the autoencoder's reconstruction capabilities. From training 210 models, we concluded that two latent variables are sufficient to capture most of the physics within the input data while selecting an optimal $\beta$ value remains complex.

The regression performance analysis shows that GPR can predict latent space coordinates with reasonable accuracy for both $\beta$-VAE and PCA+$\beta$-VAE models, with PCA pre-processing slightly improving results. The regressor performs equally well for training and testing datasets, with minimal effect from $\beta$. When combined with the decoding process, the evaluation of surrogate models in predicting wing pressure distributions confirms that both models achieve comparable global metrics. However, PCA pre-processing reduces localized errors, particularly in critical regions such as the leading edge and shockwave areas. These surrogate frameworks outperform a direct MLP model for $\beta<0.008$ at a fraction of the computational cost and include physical insight from the latent representation of the data. Hence, this surrogate approach provides a robust method for aerodynamic data prediction and highlights the importance of parameter optimization for optimal results.

To further reduce predicting errors in the surrogate model, we propose a novel strategy to fine-tune the decoder neural network. This strategy involves retraining the previously fixed decoder using the latent spaces predicted by the GPR model, allowing the decoder to learn and compensate for regression errors and further improving the $\beta$-VAE reconstruction capabilities constrained by $\beta$ regularization. The fine-tuning process significantly enhances surrogate model performance, improving accuracy and reducing dependency on the $\beta$ parameter. This adjustment allows for better generalization and performance across different flight conditions, with fine-tuned models exhibiting superior metrics, such as lower MAE and RMSE values, and more consistent performance. The fine-tuning approach enhances prediction accuracy and provides flexibility in choosing $\beta$ values for specific applications, ensuring that the surrogate model remains robust and versatile.

Overall, this research underscores the effectiveness of using $\beta$-VAEs for aerodynamic surrogate modeling. The structured latent space, robust regression performance, and the significant improvements from the fine-tuning process collectively create a highly accurate and efficient surrogate model. This model can reliably predict pressure distributions across various flight conditions, making it a valuable tool for aerodynamic analysis and optimization. The flexibility introduced by the fine-tuning process allows for tailored applications, balancing the need for interpretability and prediction accuracy, ultimately advancing the field of aerodynamic surrogate modeling.

\section*{Acknowledgments}
This work has been supported by the TIFON project, ref. PLEC2023-010251/MCIN/AEI/ 10.13039/501100011033, funded by the Spanish State Research Agency, and by AIRBUS Defence \& Space through the CETACEO project, ref. PTAG-20231008, funded by the CDTI. The authors would like to thank AIRBUS for providing the XRF1 database.

\section*{Author Declarations}
\subsection*{Declaration of Competing Interest}
The authors declare no competing financial interests or personal relationships that could have appeared to influence the work reported in this paper.
 
\subsection*{AI-Statement}
During the preparation of this work, the authors used Chat-GPT and Grammarly to improve the readability and language of this manuscript. After using this tool/service, the authors reviewed and edited the content as needed and took full responsibility for the publication's content.

\subsection*{Credit authorship contribution statement}
\textbf{V\'ictor Franc\'es-Belda}: Data curation; Formal Analysis; Methodology; Software; Visualization; Writing – original draft; Writing – review \& editing.
\textbf{Alberto Solera-Rico}: Data curation; Formal Analysis; Methodology; Software; Writing – original draft; Writing – review \& editing. 
\textbf{Javier Nieto-Centenero}: Formal Analysis; Methodology; Software; Writing – original draft; Writing – review \& editing. 
\textbf{Esther Andr\'es}:  Funding Acquisition; Resources; Project Administration;  Supervision; Validation; Writing – review \&  editing. 
\textbf{Carlos Sanmiguel Vila}: Conceptualization; Formal Analysis; Funding acquisition; Methodology;  Project Administration; Resources;  Supervision; Validation;   Writing – original draft; Writing – review \& editing. 
\textbf{Rodrigo Castellanos}: Conceptualization; Formal Analysis; Visualization; Funding acquisition; Methodology;  Project Administration; Supervision; Validation; Writing – original draft; Writing – review \& editing

\section*{Data Availability}
The dataset is subjected to the intellectual property of AIRBUS and cannot be shared, unless previous approval. The software for the surrogate has been primarily developed using PyTorch and scikit-learn frameworks. The codes used for this work are available on GitHub at \url{https://github.com/FluidMechanicsInta/Towards-aerodynamic-surrogate-modeling-based-on-beta-variational-autoencoders}.

\bibliographystyle{model1-num-names}
\bibliography{bib_v2.bib}

\end{document}